\documentclass[11pt]{article}
\usepackage{acl2015}
\usepackage{times}
\usepackage{url}
\usepackage{latexsym}
\usepackage{amsmath}
\usepackage{graphicx}
\usepackage{multirow} % For merging rows and aligning text vertically
\usepackage{makecell}
\usepackage{array}
\usepackage{placeins}
\usepackage{stfloats}
\usepackage{cuted}
\usepackage{caption} % better caption control
\usepackage{lipsum} % optional, for filler text
\usepackage[round]{natbib}
\usepackage{fvextra}
\usepackage{changepage}
\usepackage[utf8]{inputenc}
\usepackage{float}

\title{Is GPT-4 Alone Sufficient for Automated Essay Scoring?: A Comparative Judgment Approach Based on Rater Cognition}

\author{Seungju Kim \\
  KNUE, South Korea \\
  {\tt sjkim021@knue.ac.kr} \\\And  
  Meounggun Jo \\
  Hoseo University, South Korea \\  
  {\tt mkjo@hoseo.edu} }
\date{}

\begin{document}
\maketitle
\begin{strip}
    \centering
    \noindent\begin{minipage}{\textwidth}
        \includegraphics[width=0.8\linewidth]{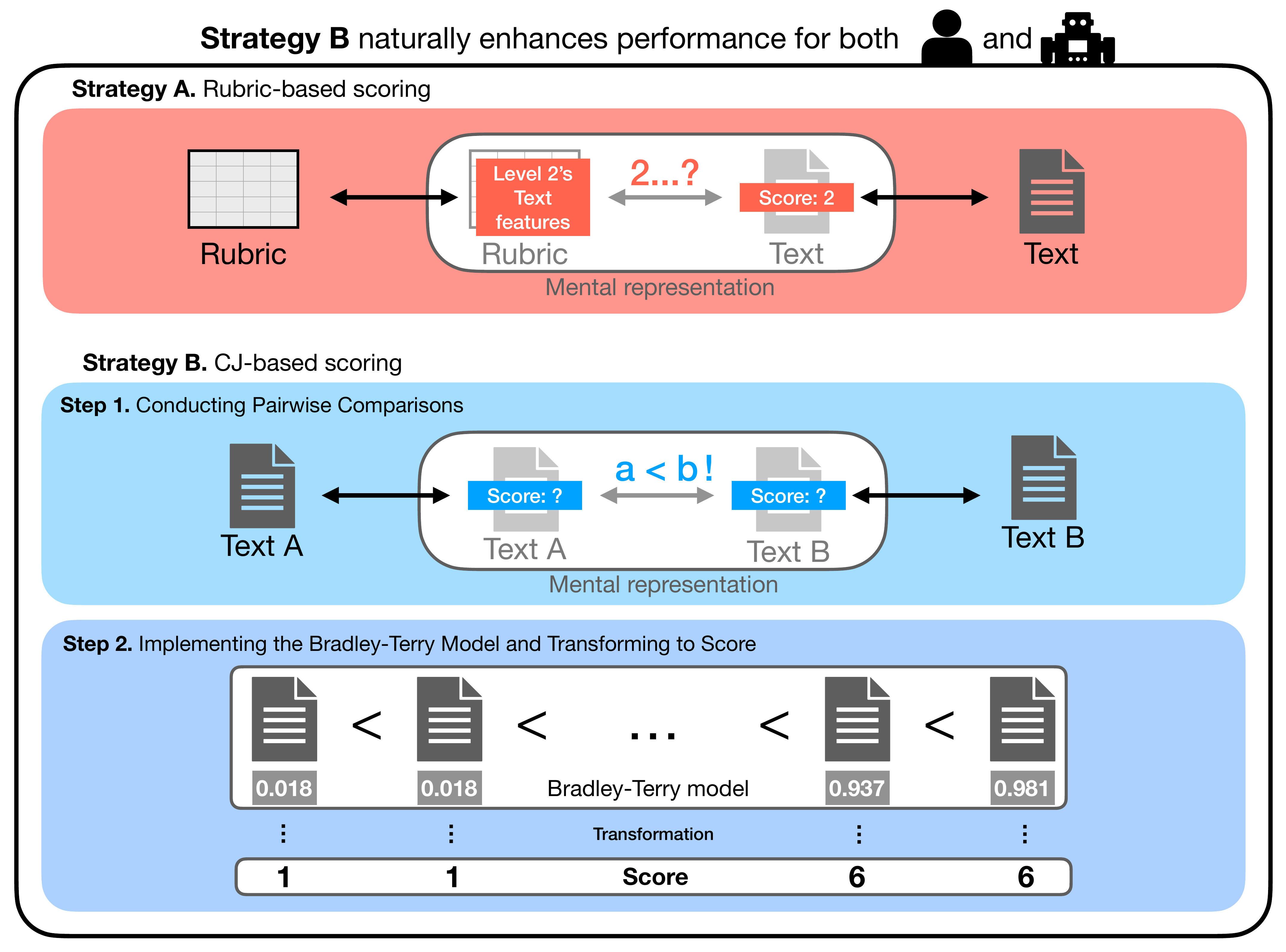}
        \centering
    \parbox{0.9\textwidth}{\centering Figure 1: Comparative overview of scoring strategies: traditional rubric-based scoring vs. two-step scoring employing comparative judgment (CJ) method} % manually creating a caption
        \label{fig:enter-label}
    \end{minipage}
\end{strip}

\begin{abstract}
  Large Language Models (LLMs) have shown promise in Automated Essay Scoring (AES), but their zero-shot and few-shot performance often falls short compared to state-of-the-art models and human raters. However, fine-tuning LLMs for each specific task is impractical due to the variety of essay prompts and rubrics used in real-world educational contexts. This study proposes a novel approach combining LLMs and Comparative Judgment (CJ) for AES, using zero-shot prompting to choose between two essays. We demonstrate that a CJ method surpasses traditional rubric-based scoring in essay scoring using LLMs.
\end{abstract}

\section{Introduction}

Essay scores are more than just numbers; they provide students with clear benchmarks for improving their writing skills and help them understand what high-quality writing looks like. Recent advancements in Large Language Models (LLMs) have shown promise in Automated Essay Scoring (AES), but their performance in zero-shot and few-shot settings often falls short compared to state-of-the-art models and human raters. While fine-tuning LLMs for specific essay scoring tasks yields better results, this approach is limited in scalability and adaptability. Especially, in real-world educational settings, diverse essay prompts and rubrics are used across various subjects, grade levels, and educational institutions. Fine-tuning LLMs for each specific task is time-consuming, resource-intensive, and impractical. Therefore, exploring zero-shot and few-shot approaches is crucial for developing AES systems that can be easily adapted to various educational settings without extensive fine-tuning.

From a psychological perspective, multi-trait essay scoring using rubrics is a cognitively demanding task for human raters \citep{Bejar:12, Hamp:91, Zhang:13}. Meanwhile, recent writing assessment research proposes Comparative Judgment (CJ) as an alternative method. CJ involves repeatedly comparing pairs of essays to produce results, offering a more cognitively intuitive approach for humans \citep{Laming:03} and highly reliable scoring results \citep{Verhavert:19}. This study starts with the question: Could the task that is natural for humans also be natural for LLMs? The combination of LLMs and CJ presents a novel approach to AES. This study investigates using few-shot prompting to enable LLMs to choose between two essays, emulating the comparative judgment process used by human raters.

\section{Related Work}

\subsection{Automated Essay Scoring}
Automated Essay Scoring (AES) is a field of research that focuses on developing computer systems to evaluate and score written essays. The goal of AES is to provide a reliable, efficient, and consistent method for assessing writing quality, which can be particularly useful in educational settings. 
\subsubsection{Performance of LLMs in AES}
Recent studies have explored the application of decoder-only Transformer-based language models, such as GPT-3.5 and GPT-4, in AES. Despite the impressive generalizability demonstrated by these models across various tasks, their potential has not been fully leveraged in the AES domain.

While fine-tuned models have shown promising results in capturing essay quality \citep{Xiao:24,Do:24}, their zero-shot and few-shot performances often fall short compared to previous state-of-the-art models \citep{Han:23,Mansour:24}. \citet{Han:23} reported that the BERT model achieved an average QWK score of 0.421, while the GPT-3.5 model with zero-shot or few-shot learning only achieved a QWK score of 0.336--0.385 on the DREsS dataset. Similarly, \citet{Mansour:24} found that on the ASAP dataset, the existing SOTA model achieved QWK scores between 0.544 and 0.771, whereas the GPT-3.5-turbo model and Llama2 model resulted in QWK scores ranging from 0.023 to 0.327. \citet{Xiao:24} also observed that GPT-4 with few-shot learning showed lower performance (0.257--0.784) compared to Fine-tuned GPT-3.5 (0.613--0.859) in all essay sets of the ASAP dataset.
\subsubsection{Limitations of Fine-tuning-based Methods}
However, fine-tuning-based methods require a large amount of data in advance, which limits their applicability in contexts where a wide variety of essay prompts and rubrics are used, except for tasks in assessment situations that are conducted in a batch manner. Furthermore, considering that essay scores are generally provided analytically rather than holistically, as mentioned by \citet{Do:24}, creating separate models or fine-tuning for each trait would require substantial resources. This suggests that in addition to fine-tuning language models, new approaches are needed to overcome the limitations of language models in extremely limited resource environments.
\subsubsection{Effects of Prompt Engineering}
Recent studies have investigated the effect of prompt engineering on the performance of LLMs in AES. \citet{Han:23} found that providing more context to GPT-3.5, particularly by requesting it to generate feedback related to the scoring rubrics, can enhance its essay scoring performance. \citet{Yancey:23} observed that GPT-4, when provided with calibration examples, can achieve a QWK close to a strong baseline AES system but lower than human raters. \citet{Mansour:24} designed four prompts with incremental improvements and found that different types of prompts yielded higher performance for different essay tasks and LLM models, with no consistent results. While prompt engineering can enhance the performance of LLMs in AES to some extent, These results highlight the need for further research and development in this area.

\subsection{Rater Cognition in Essay Scoring}

\subsubsection{Rubric-based Scoring}
Rubric-based scoring, which underlies AES, is a cognitively demanding task for human raters. The process of scoring written texts involves a complex interplay between the scorer's internal standards and external scoring rubrics, resulting in the formation of mental representations  \citep{Freedman:83,Lumley:02,Wolfe:94}. However, raters often struggle to internalize the externally provided scoring criteria \citep{Lumley:02}, which can further complicate the scoring process. While analytical scoring requires raters to assess multiple aspects of writing based on detailed criteria, this process is cognitively demanding \citep{Bejar:12} and can lead to inconsistencies in scoring outcomes due to various cognitive biases \citep{Tavares:13,Zhang:13}. Therefore, obtaining reliable scores through rubric-based scoring requires a significant investment of resources, including the development of assessment criteria and extensive training of human raters \citep{Mccaffrey:22,North:03}.

\subsubsection{Comparative Judgment}
Comparative Judgment (CJ) has been proposed as an alternative method to address the limitations of rubric-based scoring \citep{Pollitt:12}. In CJ, raters select which of two different objects (i.e., essays) is better, and by repeating this process multiple times, the rank and strength of each essay can be calculated. The concept of CJ was first introduced by \citet{Thurstone:27}, and the Bradley-Terry model \citep{Bradley:52} is commonly used to analyze the data. CJ offers a more intuitive decision-making process for raters \citep{Laming:03} and has been shown to produce highly reliable scoring results \citep{Verhavert:19}. As a result, it is considered a promising alternative to rubric-based scoring.

However, the efficiency of CJ becomes limited as the number of essays increases due to the rapidly growing number of pairs that need to be compared by human raters \citep{Bouwer:24,Goossens:18}. This scalability issue poses a significant challenge for the widespread adoption of CJ in large-scale assessment contexts. Therefore, there is a need for innovative approaches that can maintain the benefits of CJ while addressing its limitations in terms of efficiency and scalability.

\section{Research Questions}

Building upon the existing knowledge in the field of AES and rater cognition, this study explores a novel approach to utilizing LLMs for AES by employing CJ. Instead of relying on rubric-based scoring, the proposed method prompts LLMs to choose the better essay between two given essays without any additional training, using only zero-shot prompting. 
The study aims to address the following research questions:
\\

\textbf{RQ1}. When using a rubric-based scoring strategy, will the GPT-4 model be able to better imitate human-rater scores compared to the GPT-3.5 model?

\textbf{RQ2.} When using a rubric-based scoring strategy, will GPT models be able to better imitate human rater's scores if an elaborated scoring rubric with descriptors is used?

\textbf{RQ3.} When using a CJ-based scoring strategy, will the GPT model be able to better imitate human rater scores compared to the rubric-based scoring strategy?

\textbf{RQ4.} When using a CJ-based scoring strategy and utilizing fine-grained scores, will GPT models be able to better imitate human rater scores?

\section{Methods}

\subsection{Dataset}

We utilized essay sets 7 and 8 from the ASAP dataset\footnote[1]{\url{https://www.kaggle.com/c/asap-aes}}, which include multiple raters' scores and analytical scoring based on 4 and 6 traits, respectively. These two prompt sets are the only ones in the ASAP dataset that provide rubric-based scores instead of a single holistic score. Prompt set 7 consists of 1,569 essays written by 7th-grade students, with an average length of 250 words. 

The essays are scored on a scale of 0-3 across four traits (ideas, organization, style, and conventions). Prompt set 8, on the other hand, comprises 723 essays written by 10th-grade students, with an average length of 650 words. These essays are scored on a scale of 1-6 across six traits (ideas and content, organization, voice, word choice, sentence fluency, and conventions). To minimize the variance arising from the ambiguity of the rubric itself (or the diversity of rubric interpretations) and to more dramatically reveal the effects of differences between scoring strategies, such as the rubric-based method and the CJ-based method, we focused on these analytically scored essay sets.

\subsection{Models}

The LLM models used for inference in this study are the GPT-3.5 model (gpt-3.5-turbo-0123) and the GPT-4 model (gpt-4-0125-preview), both developed by OpenAI. The models were accessed through API calls, and the temperature parameter was set to 0 for all experiments.

\subsection{Rubric-based Scoring Strategy}

\subsubsection{Basic-type Rubric}
The first condition, the Basic rubric, involves the LLM scoring essays using the rubrics that were used for grading Essay Set 7 and Set 8 in the ASAP dataset. The basic rubric consists of 4 traits for Set 7 and 6 traits for Set 8, with each score having a corresponding descriptor. The descriptors for the rubric used in Set 7 are relatively simple, while those in Set 8 are very specific. The average word count per trait for the descriptors is M=66.2 (SD=15.2) for Set 7 and M=543.2 (SD=62.4) for Set 8.

\subsubsection{Elaborated-type Rubric}  
To examine the influence of rubric type on LLM performance in automated essay scoring, the rubric descriptors for Essay Set 7 were elaborated using two main methods: by adding general descriptions (Elaborated with General Description; EGD) or by including explanations of the logic between the scores awarded on the example essays and the rubric (Elaborated with Specific Examples; ESE). The gpt-4-0125-preview model was employed to generate these elaborated rubrics. The prompts used for this purpose and examples of the EGD rubric and ESE rubric are provided in Appendix A and C, respectively. The rubrics generated by the GPT-4 model were used without any modifications.

The example essays used for elaborating the rubrics were randomly sampled (seed=1, 2) from the remaining data after excluding the evaluation dataset. For each score level, a maximum of three essays that received identical scores from both raters 1 and 2 were selected. The same seed values used for extracting the evaluation dataset were employed for random sampling. However, in some cases, there were insufficient essays for certain score levels. In such instances, the rubric was elaborated by inferring from the examples of other score levels and the existing descriptions. When using either the EGD-type or ESE-type rubrics, all the available example essays were included in the prompts.

\subsection{CJ-based Scoring Strategy}

The CJ-based Scoring strategy involves choosing the better essay between two essays. The essay judged as better written in a pairwise comparison is assigned 1 if it wins, while the essay deemed relatively poor is assigned 0 if it loses. And then estimating a value representing the relative superiority of the essays through the Bradley-Terry model \citep{Bradley:52}, as shown in the equation below:
\begin{equation}
\scalebox{0.99}[1]{\(prob(A\ beats\ B\ |\lambda_{a},\ \lambda_{b}) = \frac{exp(\lambda_{a} - \lambda_{b})}{1 + exp(\lambda_{a} - \lambda_{b})}\)}
\end{equation}

In the equation above, $\lambda$ represents the quality parameter of each essay. Rasch \citep{Rasch:60} demonstrated that the optimal solution can be obtained through Maximum Likelihood (ML) estimation. For this purpose, the btm module implemented in the sirt package (v.4.1-15) in R was used (ignore.ties=True, maxiter=200).

\subsection{Evaluation}

\subsubsection{Test Data}
The number of essays used for testing varied by trait, with 31-35 essays for Set 7 and 27-31 essays for Set 8. These essays were obtained through stratified sampling from each dataset, using the average scores assigned by raters as labels. Due to insufficient data for some labels, there were differences in the number of essays tested for each trait.

For Set 7, approximately 5 essays were randomly sampled for each label, while for Set 8, around 2 essays were sampled per label. The random seeds used for sampling were fixed at 1 and 2. The number of essays used for testing was limited to manage API call costs during the comparative judgment process. The number of essays sampled for each essay set, trait, and score level is provided in Appendix D.

\subsubsection{Evaluation Method}
To evaluate the AES performance, we used Quadratic Weighted Kappa (QWK) \citep{Cohen:68}, the most widely used metric in the AES task. In the rubric-based scoring condition, scores were predicted on the same scale as the essay set's rubric, compared with each rater's scores using QWK, and then averaged. In the CJ-based scoring condition, the relatively estimated scores were converted to absolute scores for comparison with raters' scores. The prompts used for rubric-based scoring and comparative judgment are provided in Appendix B.
\begin{equation}
\scalebox{0.83}[1]{\( T(p) = R(p) \times (\max(S_{c_j}) - \min(S_{c_j})) + \min(S_{c_j}) \)}
\end{equation}
\begin{equation}
R(x) = \arg\min_{s \in S_{cj}}|s - x|
\end{equation}

The CJ results were linearly transformed to the scoring scale range of each essay set using the transformation equation (2), where equation (3) is the rounding function. The scale range $S_{cj}$ is \{0,\ 1,\ 2,\ 3\} for Essay Set 7 and \{1,\ 2,\ 3,\ 4,\ 5,\ 6\} for Essay Set 8. The finer-grained scale $S_{cjf}$ represents scores obtained by averaging raters' scores and rounding to the second decimal place, with ranges \{0.0, 0.5, 1.0, 1.5, 2.0, 2.5, 3.0\} for Essay Set 7 and \{1.0, 1.5, 2.0, 2.3, 2.5, 2.7, 3.0, 3.3, 3.5, 3.7, 4.0, 4.3, 4.5, 4.7, 5.0, 5.5, 6.0\} for Essay Set 8.

\section{Results}

\subsection{RQ1: Rubric-based Scoring with Basic-type Rubric}

As shown in Table 1, the GPT-4 model demonstrated substantially better performance compared to GPT-3.5, except for traits 5 and 6 of Essay Set 8, where performance decreased. A Wilcoxon signed-rank test revealed that the differences between the two models were statistically significant (p-value$<$.000, statistic=145).

However, despite the overall superiority of GPT-4, the traits in Essay Set 7 exhibited lower average performance compared to those in Set 8, as evident in Table 1. Specifically, for GPT-4, the QWK values ranged from 0.267 to 0.557 in Essay Set 7, while they were higher in Essay Set 8, ranging from 0.722 to 0.802.

\begin{table*}[b]
  \caption{QWK Performance Comparison: Rubric-based vs CJ-based Scoring}
  \resizebox{\textwidth}{!}{
\begin{tabular}{ccc|c|cccc|cccccc}
\hline
\multirow{2}{*}{\begin{tabular}[c]{@{}c@{}}Evaluation \\ Strategy\end{tabular}} & \multirow{2}{*}{\begin{tabular}[c]{@{}c@{}}Rubric \\ Type\end{tabular}} & \multirow{2}{*}{Model} & \multirow{2}{*}{Total} & \multicolumn{4}{c|}{Essay Set \#7} & \multicolumn{6}{c}{Essay Set \#8} \\ \cline{5-14} 
 &  &  &  & Trait1 & Trait2 & Trait3 & Trait4 & Trait1 & Trait2 & Trait3 & Trait4 & Trait5 & Trait6 \\ \hline
\multirow{2}{*}{R} & \multirow{2}{*}{B} & \multirow{2}{*}{Human} & 0.734 & \textbf{0.763} & \textbf{0.775} & 0.682 & \textbf{0.746} & 0.75 & 0.779 & 0.721 & 0.683 & 0.661 & 0.761 \\
 &  &  & (±0.073) & \textbf{±0.063} & \textbf{±0.014} & (±0.082) & \textbf{±0.055} & (±0.098) & (±0.048) & (±0.105) & (±0.124) & (±0.084) & (±0.101) \\ \hline
\multirow{2}{*}{R} & \multirow{2}{*}{B} & \multirow{2}{*}{GPT-3.5} & 0.438 & 0.399 & 0.191 & 0.271 & 0.19 & 0.532 & 0.47 & 0.633 & 0.608 & 0.734 & 0.704 \\
 &  &  & (±0.100) & (±0.131) & (±0.033) & (±0.102) & (±0.172) & (±0.090) & (±0.070) & (±0.146) & (±0.097) & (±0.065) & (±0.073) \\
\multirow{2}{*}{R} & \multirow{2}{*}{B} & \multirow{2}{*}{GPT-4} & 0.567 & 0.566 & 0.454 & 0.322 & 0.269 & 0.763 & 0.741 & 0.743 & 0.749 & 0.704 & 0.686 \\
 &  &  & (±0.102) & (±0.120) & (±0.054) & (±0.133) & (±0.083) & (±0.118) & (±0.075) & (±0.082) & (±0.072) & (±0.143) & (±0.152) \\ \hline
\multirow{2}{*}{CJ} & \multirow{2}{*}{B} & \multirow{2}{*}{GPT-3.5} & 0.573 & 0.545 & 0.437 & 0.366 & 0.506 & 0.632 & 0.738 & 0.67 & 0.739 & 0.671 & 0.648 \\
 &  &  & (±0.086) & (±0.100) & (±0.073) & (±0.029) & (±0.092) & (±0.121) & (±0.092) & (±0.091) & (±0.099) & (±0.090) & (±0.100) \\
\multirow{2}{*}{CJ} & \multirow{2}{*}{B} & \multirow{2}{*}{GPT-4} & 0.674 & 0.635 & 0.606 & 0.595 & 0.59 & 0.724 & 0.784 & 0.731 & 0.786 & 0.751 & 0.672 \\
 &  &  & (±0.087) & (±0.095) & (±0.104) & (±0.054) & (±0.059) & (±0.146) & (±0.051) & (±0.093) & (±0.068) & (±0.106) & (±0.114) \\
\multirow{2}{*}{CJ\_F} & \multirow{2}{*}{B} & \multirow{2}{*}{GPT-3.5} & 0.641 & 0.577 & 0.44 & 0.455 & 0.562 & 0.751 & 0.754 & 0.771 & 0.822 & 0.797 & 0.747 \\
 &  &  & (±0.064) & (±0.097) & (±0.018) & (±0.069) & (±0.048) & (±0.075) & (±0.121) & (±0.029) & (±0.101) & (±0.062) & (±0.037) \\
\multirow{2}{*}{CJ\_F} & \multirow{2}{*}{B} & \multirow{2}{*}{GPT-4} & \textbf{0.776} & 0.75 & 0.68 & \textbf{0.733} & 0.679 & \textbf{0.847} & \textbf{0.819} & \textbf{0.847} & \textbf{0.869} & \textbf{0.86} & \textbf{0.813} \\
 &  &  & \textbf{±0.071} & (±0.148) & (±0.092) & \textbf{±0.045} & (±0.060) & \textbf{±0.074} & \textbf{±0.088} & \textbf{±0.046} & \textbf{±0.046} & \textbf{±0.044} & \textbf{±0.038} \\ \hline
\end{tabular}
}
\end{table*}

\subsection{RQ2: Rubric-based Scoring with Elaborated-type Rubric}

In this section, we examined the impact of using elaborated rubrics with descriptors on the performance of GPT models in imitating human rater's scores for Essay Set 7. As shown in Table 2, when using the GPT-3.5 model, we observed an increase in the average QWK values across traits compared to the Basic-type (B) rubric. 

However, under the GPT-4 model condition, some traits exhibited either no difference or even a decrease in QWK values. A Wilcoxon signed-rank test revealed that the only statistically significant difference was found when using the ESE-type rubric compared to the B-type rubric with the GPT-3.5 model (p-value $<$.000, statistic=3).

\begin{table}[h]
\caption{Performance comparison of GPT models using basic and elaborated type rubrics}
\centering
\resizebox{\columnwidth}{!}{
\begin{tabular}{cc|c|cccc}
\hline
Model & \begin{tabular}[c]{@{}c@{}}Rubric\\ Type\end{tabular} & Total & Trait1 & Trait2 & Trait3 & Trait4 \\ \hline
\multirow{2}{*}{Human} & \multirow{2}{*}{B} & 0.741 & 0.763 & 0.775 & 0.682 & 0.746 \\
 &  & (±0.054) & (±0.063) & (±0.014) & (±0.082) & (±0.055) \\ \hline
\multirow{2}{*}{GPT-3.5} & \multirow{2}{*}{B} & 0.263 & 0.399 & 0.191 & 0.271 & 0.19 \\
 &  & (±0.109) & (±0.131) & (±0.033) & (±0.102) & (±0.172) \\
\multirow{2}{*}{GPT-3.5} & \multirow{2}{*}{EGD} & 0.449 & 0.637 & 0.375 & 0.464 & 0.318 \\
 &  & (±0.119) & (±0.107) & (±0.077) & (±0.049) & (±0.240) \\
\multirow{2}{*}{GPT-3.5} & \multirow{2}{*}{ESE} & 0.446 & 0.642 & 0.419 & 0.452 & 0.273 \\
 &  & (±0.053) & (±0.082) & (±0.035) & (±0.048) & (±0.047) \\ \hline
\multirow{2}{*}{GPT-4} & \multirow{2}{*}{B} & 0.403 & 0.566 & 0.454 & 0.322 & 0.269 \\
 &  & (±0.098) & (±0.120) & (±0.054) & (±0.133) & (±0.083) \\
\multirow{2}{*}{GPT-4} & \multirow{2}{*}{EGD} & 0.4 & 0.562 & 0.554 & 0.217 & 0.267 \\
 &  & (±0.067) & (±0.092) & (±0.038) & (±0.023) & (±0.115) \\
\multirow{2}{*}{GPT-4} & \multirow{2}{*}{ESE} & 0.435 & 0.566 & 0.509 & 0.367 & 0.296 \\
 &  & (±0.059) & (±0.012) & (±0.019) & (±0.071) & (±0.133) \\ \hline
\end{tabular}
}
\end{table}

\subsection{RQ3: CJ-based Scoring}

We examined the effectiveness of the CJ-based scoring strategy compared to the rubric-based scoring strategy in enabling GPT models to better imitate human rater scores. As presented in Table 1, under the CJ-based scoring condition, the average QWK values were 0.573 for GPT-3.5 and 0.674 for GPT-4, representing performance improvements of approximately 30.8\% and 18.9\%, respectively, compared to the Basic-type rubric-based scoring condition. 

Furthermore, as shown in Figure 2, a Wilcoxon signed-rank test revealed that the performance enhancements due to CJ were statistically significant, regardless of the model employed (GPT-3.5: p-value$<$.000, statistic=1092; GPT-4: p-value$<$.000, statistic=371). 

\setcounter{figure}{1}
\begin{figure}[h]
    \centering
    \includegraphics[width=1\linewidth]{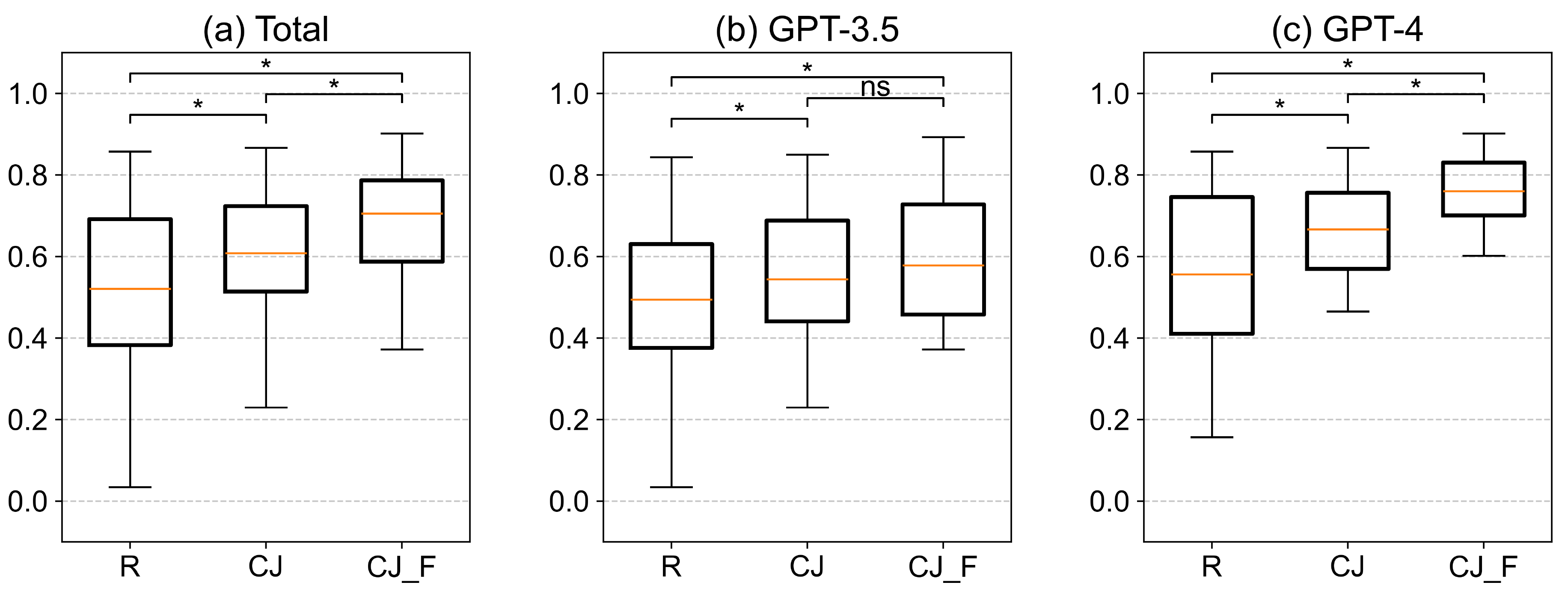}
    \caption{Performance Improvements with CJ-based Scoring Across Models}
    \label{fig:enter-label}
\end{figure}

\subsection{RQ4: CJ-based Scoring with Fine-grained Scores}

As shown in Table 1, under the fine-grained score condition (CJ\_F), both GPT models demonstrated additional performance improvements compared to the CJ condition. A Mann-Whitney U test revealed that these differences were statistically significant for the GPT-4 model (p-value$<$.000, statistic=1430). These findings suggest that incorporating fine-grained scores when using the CJ-based scoring strategy can enhance the performance of GPT models, particularly GPT-4, in imitating human rater scores. 

\section{Further Analysis}

\subsection{CJ with Elaborated Rubrics}

We further investigated the impact of using elaborated scoring rubrics in conjunction with CJ on model performance, particularly for Essay Set 7, where the initial scoring rubric was less detailed. While the overall performance was lower in the CJ condition with the basic rubric for this essay set, we aimed to determine if employing an elaborated rubric would lead to performance improvements.

As presented in Table 3, our findings suggest that using an elaborated rubric in the CJ condition resulted in some observable improvements in average scores. However, these differences were not statistically significant. 

\begin{table}[]
\caption{Performance improvements of CJ and CJ\_F across rubric types}
\centering
\resizebox{\columnwidth}{!}{
\begin{tabular}{cccc}
\hline
Model & \begin{tabular}[c]{@{}c@{}}Evaluation \\ Strategy\end{tabular} & \begin{tabular}[c]{@{}c@{}}Rubric \\ Type\end{tabular} & Total \\ \hline
Human & R & B & 0.741(±0.059) \\ \hline
GPT-3.5 & CJ & B & 0.464(±0.099) \\
GPT-3.5 & CJ & EGD & 0.446(±0.114) \\
GPT-3.5 & CJ & ESE & 0.449(±0.094) \\ \hline
GPT-3.5 & CJ\_F & B & 0.508(±0.082) \\
GPT-3.5 & CJ\_F & EGD & 0.502(±0.107) \\
GPT-3.5 & CJ\_F & ESE & 0.519(±0.103) \\ \hline
GPT-4 & CJ & B & 0.607(±0.075) \\
GPT-4 & CJ & EGD & 0.602(±0.073) \\
GPT-4 & CJ & ESE & 0.624(±0.064) \\ \hline
GPT-4 & CJ\_F & B & 0.710(±0.079) \\
GPT-4 & CJ\_F & EGD & 0.712(±0.063) \\
GPT-4 & CJ\_F & ESE & 0.726(±0.088) \\ \hline
\end{tabular}
}
\end{table}

\subsection{Effectiveness of CJ-based approach between rubric types}

To further examine whether the effects of the CJ and CJ\_F conditions were statistically significant across different rubric types, we conducted a Wilcoxon signed-rank test. As illustrated in Figure 2, the results showed that the performance improvements from the R condition to the CJ and CJ\_F condition were statistically significant, regardless of the rubric type.

\begin{figure}[h]
    \centering
    \includegraphics[width=1\linewidth]{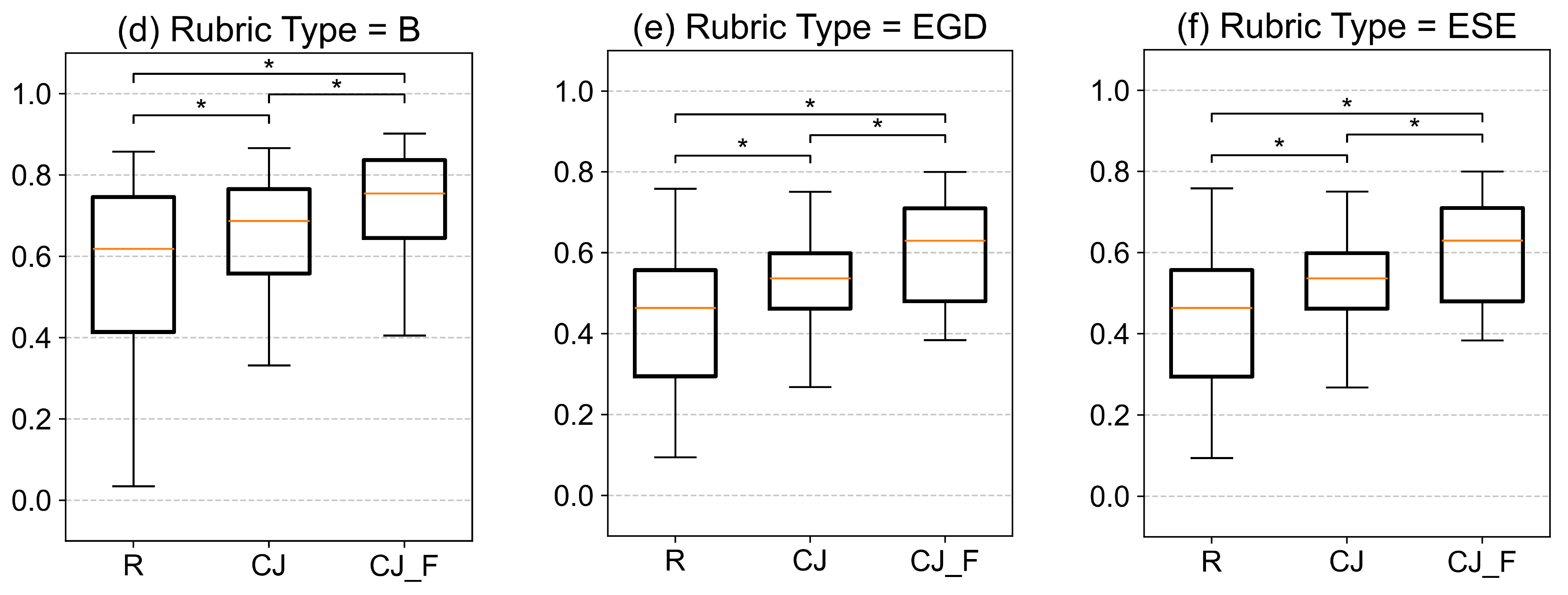}
    \caption{Performance Improvements of CJ and CJ\_F Across Rubric Types}
    \label{fig:enter-label}
\end{figure}

\section{Discussion}

This research illustrates the potential use of Large Language Models (LLMs) with Comparative Judgment (CJ) for Automated Essay Scoring (AES). The results provide valuable insights into how LLMs can be effectively utilized in this area. In the following discussion, we will closely examine these findings and analyze their significance for the field of AES. 

\subsection{Impact of Essay Set Characteristics on LLM Performance}
The present study highlights the substantial impact of essay set characteristics on the performance of LLMs, even for the advanced GPT-4 model. A marked disparity was observed between essay sets 7 and 8, suggesting that factors beyond the model's inherent capabilities, such as the level of detail in scoring rubrics, play a pivotal role in determining AES performance.

The lack of specificity in the rubrics for essay set 7, which contained approximately nine times fewer words per sub-trait compared to set 8, likely led the LLM to evaluate set 7 based on logic and evidence that diverged from human raters. These findings underscore the importance of providing comprehensive and well-defined scoring criteria to guide the judgment of LLMs in AES tasks.

Traits 4 of essay set 7 and 6 of essay set 8, both related to the evaluation criteria for conventions, exemplify this divergence. LLMs demonstrated a capacity for rigorous analysis of error characteristics and nuances, focusing intently on detailed aspects of the text. In contrast, human raters may apply these evaluation criteria from a more qualitative perspective, such as whether the level of errors interferes with their understanding of the text content \citep{Cumming:02}. Further research is needed to better understand and address these discrepancies between LLM and human rater judgments.

\subsection{Influence of Elaborated Scoring Rubrics on GPT Models}
The study reveals the varying impact of elaborated scoring rubrics on the performance of different GPT models. While GPT-3.5 generally benefited from more detailed rubrics, GPT-4 exhibited mixed results, with some traits even showing a decrease in performance. This suggests that the rubrics may have been overfitted to the essay dataset used in the elaboration process.

It is noteworthy that for essay set 8, using GPT-4 with basic-type rubrics alone yielded QWK scores ranging from 0.686 to 0.763. These values are similar to the performance level achieved when using GPT-4 with basic-type rubrics in the CJ condition, highlighting the importance of elaborated rubrics. Furthermore, changes in rubrics influence performance improvements or deterioration even under conditions utilizing the CJ strategy, demonstrating that rubrics remain an important factor in essay evaluation.

However, considering that the general rubric development process is iterative and resource-demanding \citep{Janssen:15,Mccaffrey:22}, further research is needed on how LLMs can effectively assist this process. Investigating methods for utilizing LLMs to create comprehensive and well-defined scoring criteria should be a priority to enhance the accuracy and efficiency of AES systems.

\subsection{Effectiveness of CJ-based Scoring Strategy}
The CJ-based scoring strategy proved to be more effective than the traditional rubric-based method in enabling GPT models to emulate human rater scores, with significant performance improvements observed for both GPT-3.5 and GPT-4. However, it is important to consider that when scoring, human raters not only compare essays but also clearly connect rubric descriptors with essay features \citep{Cumming:02}. This approach may sometimes be more economical than comparing multiple pairs of essays.

For future research, adopting a two-way approach that reflects these human cognitive processes by utilizing both methods appears promising in terms of both efficiency and reliability.

\subsection{Utilizing Fine-grained Scores in CJ-based Scoring}
Incorporating fine-grained scores in the CJ-based scoring strategy further augmented the performance of GPT models, particularly GPT-4. This finding underscores the value of utilizing granular scoring information to improve the accuracy of AES systems powered by advanced language models.

Generally, scores and scoring results are referred to as "score bands," which represent categories of ability levels that exist on a continuous scale. However, these scores are given as discrete values, which means that machines have no choice but to understand these values discretely, and the scores can be distorted depending on how we assign tasks to LLMs. It is important to consider that human perception of writing quality is more nuanced and granular. As such, the development of datasets constructed using the CJ approach from the outset could enable a more rigorous validation of LLM judgments and align more closely with human intuition.

\section{Future Work}

This study has demonstrated the potential of combining Large Language Models (LLMs) with Comparative Judgment (CJ) for Automated Essay Scoring (AES). However, there are several avenues for future research that can further enhance the generalizability, robustness, and practical applicability of this approach.

\subsection{Validation on Diverse Datasets}

While CJ proved to be the most effective strategy in this study, the performance varied considerably depending on the trait and essay set. As the writing tasks in Essay Sets 7 and 8 were narrative in nature, it is necessary to verify whether this approach can effectively work on data from other types of writing tasks, such as persuasive writing. Currently, there is a lack of publicly available rubric-based evaluation datasets to test this. Although some datasets, such as ASAP++ \citep{Mathias:18}, provide scores based on specific traits, it is unclear which rubrics were used to score these traits. However, with the recent release of publicly accessible rubric-based evaluation datasets like DREsS \citep{Yoo:24}, further validation on various datasets is necessary.

\subsection{Assigning Absolute Scores}

This study assumed a uniform distribution and employed Bradley-Terry modeling and linear transformation of CJ estimates. However, methods for assigning absolute scores to essays require further development. In scenarios with imbalanced data, the use of the Bradley-Terry model may lead to bias in parameter estimation. Additionally, the Maximum Likelihood Estimation (MLE) process utilized in this research could potentially face nonconvergence issues.

While mathematical and statistical methods and alternatives exist to address these challenges, data augmentation methods show particular promise. It is established that certain textual factors can complicate essay evaluation \citep{Wolfe:16}, suggesting that human raters may find it especially difficult to judge texts with specific characteristics. Hypothetically, if this principle extends to LLMs, incorporating generated data points with features conducive to easier evaluation into the essay set might yield additional performance improvements.

Furthermore, by having human raters pre-judge the absolute grades of some generated data that possess features conducive to easier evaluation, these could serve as model essays for LLM evaluation. If LLMs then assess the remaining essays through comparison with these model essays, the resulting scores may transcend mere relative rankings and carry absolute meaning.

\subsection{Human-AI Interaction}

In educational settings where resources are limited and fine-tuning is not feasible, it is crucial to investigate how these technologies can effectively support assessment while collaborating with teachers. Human raters are susceptible to cognitive biases, and even the evaluation data used in this study, despite its extensive use in previous research, may not be entirely error-free or of the highest quality. Had LLMs assisted human raters in creating the evaluation data from the outset, this study's results might have differed slightly.

Although this study successfully enabled LLMs to perform evaluations similar to humans by modifying their operational method without separate fine-tuning, the full automation approach has limitations in supporting human evaluators' reading or assessment processes. In scenarios requiring human-AI interaction, it is crucial for LLMs to be finely adjustable (controllable) and sufficiently interpretable. This aspect warrants further research to enhance the synergy between human expertise and AI capabilities in educational assessment.

\subsection{Optimization of Comparison Pairs}

Due to the cost limitations of using the GPT-4 model, this study could not validate the approach on larger datasets. This is partly due to the problem of the number of pairs to be compared increasing exponentially when applying CJ \citep{Goossens:18}. Future research should integrate methods such as Adaptive Comparative Judgment (ACJ) \citep{Pollitt:12} to optimize the number of comparison pairs and verify the effectiveness of such approaches.

\section{Conclusion}

This study contributes to the growing body of research on the application of Large Language Models (LLMs) in Automated Essay Scoring (AES) by investigating the effectiveness of combining LLMs with Comparative Judgment (CJ). The findings demonstrate that the CJ-based scoring strategy, particularly when combined with elaborated rubrics and fine-grained scores using GPT-4, is more effective than the traditional rubric-based scoring in enabling LLMs to imitate human rater scores. This study shows that while GPT-4 is a powerful tool for AES, it is not sufficient on its own, as many factors influence both human raters and LLMs in essay scoring.

The results have significant implications not just for the advancement and utilization of LLMs in AES, but also for several research domains that entail generating multi-trait scoring data with a hierarchy. The insight gained from this study can guide the development of automated scoring systems in various fields, emphasizing the significance of taking into account elements such as scoring criteria, scoring methods, and the specific language model used. This work highlights the significance of interdisciplinary collaboration among specialists in the areas of natural language processing, educational assessment, and cognitive psychology to further enhance the progress and implementation of LLMs in intricate educational problems.

\bibliographystyle{plainnat}
\bibliography{references}

\newpage
\onecolumn
\appendix
\section*{Appendix}

\section{Prompts for Rubric Elaboration}
\begin{table}[H]
\centering
\begin{tabular}{|p{0.9\columnwidth}|}
\hline
{\ttfamily\spaceskip=\fontdimen2\font plus 3pt minus 2pt
Below are representative essay examples for each score on the "\{criteria\_name\}" aspect of the essay grading scale. Use the essay examples to elaborate on existing descriptors. Create specific descriptors for each score, but write them as generalised statements.

//Grading scale: \{scale\_to\_elaborate\}

//Example essay:

- Score 3: 

//Essay1: \{essay\#1\_content\}

//Essay2: \{essay\#2\_content\}

//Essay3: \{essay\#3\_content\}

…

//Writing task: \{essay\_prompt\}
}
\\ \hline
\end{tabular}
\caption*{Prompt for elaboration (EGD-type)}
\label{tab:cj_prompt}
\end{table}
\begin{table}[H]
\centering
\begin{tabular}{|p{0.9\columnwidth}|}
\hline
{\ttfamily\spaceskip=\fontdimen2\font plus 3pt minus 2pt
Below are representative essay examples for each score on the "\{criteria\_name\}" aspect of the essay grading scale. Use the essay examples to elaborate on existing descriptors. Elaborate descriptors for each score, with specific examples.

//Grading scale: \{scale\_to\_elaborate\}

//Example essay:

- Score 3: 

//Essay1: \{essay\#1\_content\}

//Essay2: \{essay\#2\_content\}

//Essay3: \{essay\#3\_content\}

…

//Writing task: \{essay\_prompt\}
}
\\ \hline
\end{tabular}
\caption*{Prompt for elaboration (ESE-type)}
\label{tab:cj_prompt}
\end{table}
\newpage
\section{Prompts for Evaluation}

\subsection{Prompt for Rubric-based Scoring}

\begin{table}[H]
\centering
\begin{tabular}{|p{0.9\columnwidth}|}
\hline
{\ttfamily\spaceskip=\fontdimen2\font plus 3pt minus 2pt
Q. Please score student writing according to the criteria given in the '\{criteria\_name\}' aspect.

//Criteria: \{criteria\}

//Answer format: \{'score\_explanation': [content], 'score': [number]\} score = [0, 1, 2, 3] Please answer only in the above dictionary format.

//Prompt: \{essay\_prompt\}

//Essay: \{essay\_content\}
}
\\ \hline
\end{tabular}
\caption*{Prompt for Scoring (B-type rubric)}
\label{tab:basic_criteria}
\end{table}

\begin{table}[H]
\centering
\begin{tabular}{|p{0.9\columnwidth}|}
\hline
{\ttfamily\spaceskip=\fontdimen2\font plus 3pt minus 2pt
Q. Please score student writing according to the scoring examples and criteria given in the '\{criteria\_name\}' aspect.

//Scoring examples: \{examples\}

//Criteria: \{elaborated criteria with general description\}

//Answer format: \{'score\_explanation': [content], 'score': [number]\} score = [0, 1, 2, 3] Please answer only in the above dictionary format.

//Prompt: \{essay\_prompt\}

//Essay: \{essay\_content\}
}
\\ \hline
\end{tabular}
\caption*{Prompt for Scoring (EGD-type or ESE-type rubric)}
\label{tab:elaborated_criteria_general}
\end{table}

\subsection{Prompt for Comparative Judgment}

\begin{table}[H]
\centering
\begin{tabular}{|p{0.9\columnwidth}|}
\hline
{\ttfamily
\spaceskip=\fontdimen2\font plus 3pt minus 2pt
Q. You're a writing assessment expert. Compare two essays (Essay A, Essay B) based on the criteria below and choose which one did better. Please answer without explanation. (e.g., Essay A or Essay B)

//Criteria:

\{criteria\_name\}

\{criteria\}

//Prompt:

\{essay\_prompt\}

Essay A: \{essayA\_content\}

//Essay B: \{essayB\_content\}
}
\\ \hline
\end{tabular}
\caption*{Prompt for Comparative Judgement}
\label{tab:elaborated_criteria_general}
\end{table}

\newpage
\section{Example of Rubric}
\begin{table}[h!]
\centering
\begin{itemize}
    \item Basic-type Rubric
\end{itemize}
\begin{tabular}{|p{0.9\columnwidth}|}
\hline
{\ttfamily\spaceskip=\fontdimen2\font plus 3pt minus 2pt
Ideas

\bigskip
Score 3: Tells a story with ideas that are clearly focused on the topic and are thoroughly developed with specific, relevant details.

Score 2: Tells a story with ideas that are somewhat focused on the topic and are developed with a mix of specific and/or general details.

Score 1: Tells a story with ideas that are minimally focused on the topic and developed with limited and/or general details.

Score 0: Ideas are not focused on the task and/or are  
undeveloped.
}
\\ \hline
\end{tabular}
\label{tab:cj_prompt}
\end{table}

\begin{table}[h!]
\centering
\begin{itemize}
    \item EGD-type Rubric
\end{itemize}
\begin{tabular}{|p{0.9\columnwidth}|}
\hline
{\ttfamily\spaceskip=\fontdimen2\font plus 3pt minus 2pt
**Ideas**

\bigskip
**Score 3:**

- The essay presents a narrative that is directly aligned with the prompt, showcasing a deep understanding of the concept of patience.

- Ideas are not only relevant but are also enriched with vivid, specific details that enhance the story, making it engaging and illustrative of the theme.

- The narrative structure is coherent, with a clear beginning, development, and conclusion that collectively underscore the significance of patience.

- The writer effectively uses descriptive language and personal reflections to convey emotions and insights, making the theme of patience resonate with the reader.

\bigskip
**Score 2:**

- The essay addresses the prompt, but the connection to the theme of patience may be less direct or slightly obscured by less relevant details.

- Ideas are generally focused on the topic of patience, but the narrative may include a mix of specific and general details, diluting the overall impact.

- The narrative demonstrates an attempt to explore the theme of patience, but the development of ideas may lack depth or fail to fully captivate the reader.

- While the essay aims to convey a message about patience, the execution may be uneven, with some parts more effectively illustrated than others.
}
\\ \hline
\end{tabular}
\label{tab:cj_prompt}
\end{table}

\begin{table}[h!]
\centering
\begin{tabular}{|p{0.9\columnwidth}|}
\hline
{\ttfamily\spaceskip=\fontdimen2\font plus 3pt minus 2pt
**Score 1:**

- The essay makes a minimal effort to address the prompt, with the theme of patience only loosely connected to the narrative.

- Ideas related to patience are underdeveloped and rely heavily on generalities, lacking the specific details necessary to bring the story to life.

- The narrative may struggle to maintain focus, with tangential or irrelevant details detracting from the theme of patience.

- The exploration of patience is superficial, offering little insight or reflection, and fails to engage the reader meaningfully.

\bigskip
**Score 0:**
- The essay fails to address the prompt, with the narrative either not focused on patience or completely lacking in development of the theme.
- Ideas are scattered or absent, with no discernible attempt to explore the concept of patience in any meaningful way.
- The narrative lacks coherence, making it difficult to identify a central theme or purpose related to patience.
- The essay does not meet the basic requirements of the task, offering no insight, reflection, or narrative related to patience.
}
\\ \hline
\end{tabular}
\label{tab:cj_prompt}
\end{table}

\begin{table}[h!]
\centering
\begin{itemize}
\bigskip
    \item ESE-type Rubric
\end{itemize}
\begin{tabular}{|p{0.9\columnwidth}|}
\hline
{\ttfamily\spaceskip=\fontdimen2\font plus 3pt minus 2pt
**Ideas**
\bigskip

**Score 3 Descriptors and Examples:**
\bigskip

- Ideas are clearly focused on the topic of patience, demonstrating a deep understanding and engagement with the concept.

- The narrative is thoroughly developed with specific, relevant details that enhance the story and provide insight into the theme of patience.

- Examples:

\hspace*{0.5cm}- Essay1 vividly describes the frustration of waiting in line for food, the anticipation, and the eventual reward of patience, culminating in receiving a warm cookie. This story effectively uses specific details like the empty food trays and the aroma of pizza to illustrate the experience of patience.
  
\hspace*{0.5cm}- Essay2 captures the excitement and patience involved in fishing, detailing the struggle with a large fish and the eventual loss, highlighting patience in the face of disappointment.
  
\hspace*{0.5cm}- Essay3 focuses on the anticipation and long wait for a ride at an amusement park, using specific details like the length of the car ride and the queue for the ride to illustrate the theme of patience and the eventual payoff of an enjoyable experience.

}
\\ \hline
\end{tabular}
\label{tab:cj_prompt}
\end{table}

\begin{table}[H]
\centering
\begin{tabular}{|p{0.9\columnwidth}|}
\hline
{\ttfamily\spaceskip=\fontdimen2\font plus 3pt minus 2pt
\hspace*{0.5cm}- Essay2 captures the excitement and patience involved in fishing, detailing the struggle with a large fish and the eventual loss, highlighting patience in the face of disappointment.
\hspace*{0.5cm}- Essay3 focuses on the anticipation and long wait for a ride at an amusement park, using specific details like the length of the car ride and the queue for the ride to illustrate the theme of patience and the eventual payoff of an enjoyable experience.

\bigskip
**Score 2 Descriptors and Examples:**

- Ideas are somewhat focused on the topic of patience but may include some irrelevant details or slightly off-topic content.
- The narrative is developed with a mix of specific and general details, which sometimes dilutes the focus or clarity of the theme of patience.

- Examples:

\hspace*{0.5cm}- Essay1 discusses the concept of patience in the context of waiting for a grade improvement, but the narrative includes a mix of specific scenarios and more general statements about patience, making the focus less clear.

\hspace*{0.5cm}- Essay2 describes the experience of shopping in a crowded store, which is relevant to patience, but the story includes some general complaints and lacks the depth of specific details that would more effectively illustrate patience.

\hspace*{0.5cm}- Essay3 recounts waiting in a long line at customs, a situation that requires patience. However, the narrative is more of a straightforward account with fewer vivid, specific details that would enrich the theme.

\bigskip
**Score 1 Descriptors and Examples:**

- Ideas are minimally focused on the topic of patience, with the narrative often veering off-topic or lacking a clear connection to the theme.

- The narrative is developed with limited and/or general details, which fails to provide a meaningful insight into the concept of patience or to engage the reader effectively.

- Examples:

\hspace*{0.5cm}- Essay1 briefly mentions hunting and fishing as activities requiring patience but offers very little detail or development, making the connection to patience weak and the narrative underdeveloped.

\hspace*{0.5cm}- Essay2 confuses the concept of being a patient in a medical sense with the theme of patience, resulting in a narrative that is off-topic and lacks focus.

\hspace*{0.5cm}- Essay3 mentions waiting at a volleyball tournament but provides minimal detail about the experience, resulting in a narrative that barely touches on the theme of patience.

\bigskip
**Score 0 Descriptors and Examples:**

- Ideas are not focused on the task of discussing patience, with narratives that are either completely off-topic or so underdeveloped that they fail to address the theme meaningfully.
}
\\ \hline
\end{tabular}
\label{tab:cj_prompt}
\end{table}

\newpage
\begin{table}[H]
\centering
\begin{tabular}{|p{0.9\columnwidth}|}
\hline
{\ttfamily\spaceskip=\fontdimen2\font plus 3pt minus 2pt
- The narrative lacks development, with no clear storyline or details related to patience, making it difficult to discern any meaningful engagement with the topic.

- Examples:

\hspace*{0.5cm}- Essay1 rambles about various situations where one might need to be patient but lacks a coherent narrative or specific details related to personal experiences of patience, making it off-topic and undeveloped.

\hspace*{0.5cm}- Essay2 makes general statements about patience without providing any narrative or examples, resulting in a piece that is undeveloped and fails to meet the task.

\hspace*{0.5cm}- Essay3 expresses a personal disinterest in patience without offering a narrative or examples, making it off-topic and not focused on the task of writing about patience.
}
\\ \hline
\end{tabular}
\label{tab:cj_prompt}
\end{table}

\section{Number of Essays Sampled for Testing}
\begin{table}[H]
\centering
\caption*{Essay set \#7}
{\small % Reducing font size
\begin{tabular}{|c|c|c|c|c|}
\hline
\textbf{Label} & Trait1 & Trait2 & Trait3 & Trait4 \\ \hline
0.0 & 5 & 2 & 1 & 2 \\ \hline
0.5 & 5 & 4 & 5 & 5 \\ \hline
1.0 & 5 & 5 & 5 & 5 \\ \hline
1.5 & 5 & 5 & 5 & 5 \\ \hline
2.0 & 5 & 5 & 5 & 5 \\ \hline
2.5 & 5 & 5 & 5 & 5 \\ \hline
3.0 & 5 & 5 & 5 & 5 \\ \hline
\textbf{Total} & 35 & 31 & 31 & 32 \\ \hline
\end{tabular}
}
\end{table}

\begin{table}[H]
\centering
\caption*{Essay set \#8}
{\small
\begin{tabular}{|c|c|c|c|c|c|c|}
\hline
\textbf{Label} & Trait1 & Trait2 & Trait3 & Trait4 & Trait5 & Trait6 \\ \hline
1.0 & 1 & 1 & 1 & 1 & 1 & 1 \\ \hline
1.5 & 1 & 1 & 0 & 1 & 1 & 1 \\ \hline
2.0 & 2 & 2 & 2 & 1 & 2 & 2 \\ \hline
2.3 & 1 & 2 & 0 & 0 & 1 & 1 \\ \hline
2.5 & 2 & 2 & 2 & 2 & 2 & 2 \\ \hline
2.7 & 1 & 2 & 1 & 1 & 2 & 2 \\ \hline
3.0 & 2 & 2 & 2 & 2 & 2 & 2 \\ \hline
3.3 & 2 & 2 & 2 & 2 & 2 & 2 \\ \hline
3.5 & 2 & 2 & 2 & 2 & 2 & 2 \\ \hline
3.7 & 2 & 2 & 2 & 2 & 2 & 2 \\ \hline
4.0 & 2 & 2 & 2 & 2 & 2 & 2 \\ \hline
4.3 & 2 & 2 & 2 & 2 & 2 & 2 \\ \hline
4.5 & 2 & 2 & 2 & 2 & 2 & 2 \\ \hline
4.7 & 2 & 2 & 2 & 2 & 2 & 2 \\ \hline
5.0 & 2 & 2 & 2 & 2 & 2 & 2 \\ \hline
5.3 & 1 & 1 & 2 & 1 & 0 & 0 \\ \hline
5.5 & 2 & 1 & 2 & 1 & 1 & 1 \\ \hline
5.7 & 1 & 0 & 1 & 0 & 0 & 0 \\ \hline
6.0 & 1 & 1 & 1 & 1 & 1 & 1 \\ \hline
\textbf{Total} & 31 & 31 & 30 & 27 & 29 & 29 \\ \hline
\end{tabular}
}
\end{table}

\end{document}